\documentclass{article}
\usepackage{spconf,amsmath,graphicx,hyperref}
\usepackage{array}
\usepackage{amssymb}
\usepackage{multirow}
\usepackage{xcolor, colortbl}
\usepackage{amsfonts, bm}

\definecolor{mygray}{gray}{.9}
\definecolor{mypink}{rgb}{.99,.91,.95}
\definecolor{mygreen}{rgb}{.9,.99,.9}

\title{The Harder The Better: Maintaining Supervised Fine-tuning Generalization with Less but Harder Data}

\name{
Zhaoyang Shang\textsuperscript{\scriptsize 1,2,\dag}, Sibo Wei\textsuperscript{\scriptsize 1,\dag}, Jianbin Guo\textsuperscript{\scriptsize 1,2,3,*,\dag}
\thanks{$^\dag$These authors contributed equally to this work.}, Rui Zhou\textsuperscript{\scriptsize 2}, Lifeng Dong\textsuperscript{\scriptsize 1}, Yin Luo\textsuperscript{\scriptsize 1}\thanks{$^*$Corresponding Author (email: jianbin.guo@wenge.com)}
}
\address{
	$^1$Beijing Wenge Technology Co.,Ltd\\
    $^2$College of Intelligence and Computing, Tianjin University\\
    $^3$Institute of Automation, Chinese Academy of Sciences
}

\begin{document}
\maketitle

\begin{abstract}
Large Language Models (LLMs) excel in general tasks, but adapting them to specialized domains relies on high-quality supervised fine-tuning (SFT) data. Although existing methods can identify subsets of high-quality data and reduce training cost to some extent, their selection process still suffers from over-reliance on LLMs’ internal knowledge, weak interpretability, and limited generalization.
To address these limitations, we propose THTB (\textbf{T}he \textbf{H}arder \textbf{T}he \textbf{B}etter), a cognitive science-inspired framework for instruction data selection and annotation guidance. THTB prioritizes higher-level cognitive instructions by combining quality filtering with intrinsic and extrinsic hardness scoring, offering interpretable and quantifiable criteria for efficient SFT, both in data selection and annotation guidance.
Experiments show that THTB enables models trained on only 5\% of the data to outperform full-dataset training, while achieving superior generalization compared with LLM-only selection. In addition, THTB provides effective annotation guidance in vertical domains, enabling a model trained on just 2\% of the data to surpass models trained on much larger datasets, demonstrating strong potential for domain adaptation. Our code, datasets, and models are available on \href{https://github.com/DYJG-research/THTB}{https://github.com/DYJG-research/THTB}.

\end{abstract}

\begin{keywords}
Large Language Model, Post-training, Data Selection
\end{keywords}

\section{Introduction}
Large Language Models (LLMs) have transformed artificial intelligence applications across diverse industries, demonstrating strong general-purpose capabilities\cite{annepaka2025large}. A key technique enabling their adaptation to specialized domains is supervised fine-tuning (SFT), for which the construction of high-quality instruction data is essential\cite{zhang2024survey}.

Early research emphasized building large-scale, diverse, and high-quality instruction datasets\cite{wang2023self, ding2023enhancing}. However, Zhou et al.\cite{zhou2023lima} showed that most of an LLM’s knowledge is acquired during pre-training, and that SFT requires only a small amount of instruction data to achieve strong performance. This insight has shifted attention toward selecting high-quality subsets from large instruction datasets\cite{chenalpagasus,ge2024clustering,li2024superfiltering}. While such methods reduce training cost and improve efficiency, they still face three limitations: (1) over-reliance on the internal knowledge of LLMs; (2) lack of concrete, intuitive quantitative metrics, resulting in low interpretability and limited guidance for annotation; (3) limited preservation of the base model’s generalization capability.

To address these issues, we propose THTB (The Harder The Better), a framework for both instruction data selection and annotation guidance. 

THTB focuses on complex, challenging instructions that enhance reasoning and generalization. This framework is grounded in Bloom’s Taxonomy~\cite{anderson2001taxonomy}, an educational classification framework introduced by Benjamin Bloom, which organizes the cognitive domain into six hierarchical levels---\textit{remember}, \textit{understand}, \textit{apply}, \textit{analyze}, \textit{evaluate}, and \textit{create}---emphasizing the progression from lower-level to higher-level (i.e., from simple to difficult) thinking skills. Based on this cognitive theory, THTB selects instruction data at higher cognitive levels (i.e., more difficult) through quality filtering, intrinsic hardness scoring, and extrinsic hardness scoring to enable efficient and generalizable SFT. Our main contributions are summarized as follows:
\begin{itemize}
    \item We introduce THTB, an automated data selection framework. Compared with prior LLM-based methods, our approach offers greater quantifiability and interpretability, and directly informs instruction data construction and annotation, reducing cost while improving effectiveness.
    \item Experiments show that, under THTB guidance, training with only 5\% of the dataset outperforms training on the full dataset. At equal data scales, THTB achieves substantially better generalization than LLM-only selection methods.
    \item We apply THTB to the Traditional Chinese Medicine domain. With THTB, a model trained on a dataset only 2\% the size of another achieves superior domain performance and generalization, offering valuable insights for other vertical domains.
\end{itemize}

\section{Methodology}
\begin{figure*}[htbp]
    \centering
    \includegraphics[width=1\textwidth]{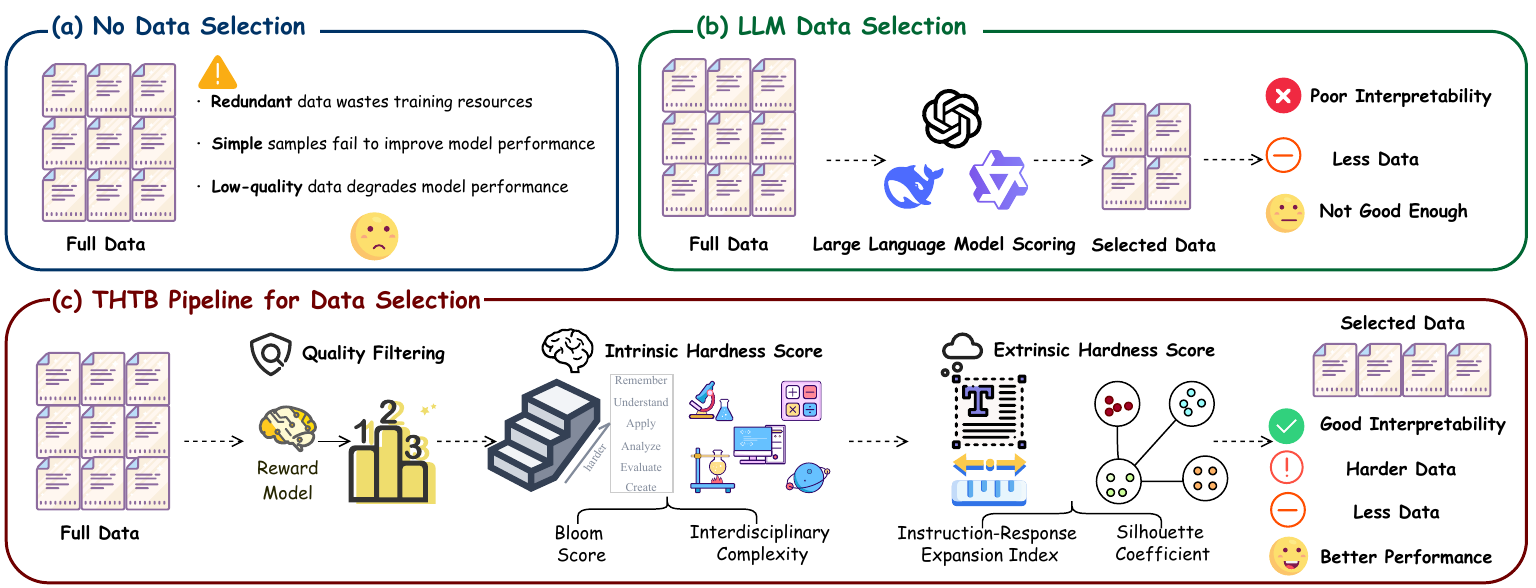} 
    \caption{Comparison of data selection methods: From no selection and LLM-based scoring to the proposed THTB pipeline.}
    \label{fig:model}
\end{figure*}

Drawing on Bloom’s taxonomy of the cognitive domain, our approach targets the selection of instruction data at higher cognitive levels—that is, data that are harder.
As shown in Figure~\ref{fig:model}, our method selects data in three stages.

\subsection{Quality Filtering}

To ensure the basic quality of the data, we first perform an initial filtering (e.g., excluding low-quality instructions or misaligned with human preferences) using a reward model~\cite{cai2024internlm2}:
\begin{equation}
    Reward = RM(inst,resp)
\end{equation}
where $inst$ and $resp$ represent the instruction and response in supervised fine-tuning, respectively. 

\subsection{Intrinsic Hardness Score}
We define the Intrinsic Hardness Score as the mean of the Bloom Score and Interdisciplinary Complexity, which characterizes the inherent difficulty of the data.
\subsubsection{Bloom Score}

Based on Bloom's Taxonomy, we introduce the Bloom score to reflect the cognitive difficulty of the data. We first use a large language model to classify each data sample into the corresponding cognitive level(s), and then compute the score using the following formula:
\begin{equation}
    Bloom = \frac{{\sum\limits_{i = 1}^6 {i \times {\delta _{{\mathbb{C}_i}}}}  - Bloo{m_{\min }}}}{{Bloo{m_{\max }} - Bloo{m_{\min }}}}
\end{equation}

where ${\delta _{{\mathbb{C}_i}}} \in \{ 0,1\} $ indicates whether the data belongs to the corresponding cognitive level. ${Bloo{m_{\min }}}$ and ${Bloo{m_{\max }}}$ represent the minimum and maximum Bloom scores in the dataset, respectively. $\mathbb{C}$ denotes the set of cognitive levels:
\begin{equation}
    \mathbb{C} = \{ {\text{Remember,Uderstand,Apply,Analyze,Evaluate,Create}}\}
\end{equation}
\subsubsection{Interdisciplinary Complexity}

We propose Interdisciplinary Complexity (IC) as a measure of the difficulty involved in solving an instruction, based on the diversity and separation of the disciplines it spans. Specifically, the more disciplines an instruction involves, and the greater the distance between these disciplines, the more challenging it is to solve.

To compute IC, we first use a large language model \cite{yang2025qwen3} to classify each data sample into one or more relevant disciplines. Next, we collect the set of all disciplines involved across the dataset. For each discipline, we prompt a large language model to generate a detailed description. We then obtain the embedding of each description using the \emph{bge-large-en-v1.5} model \cite{xiao2024c}.

For a given data sample $t$, let the associated disciplines be denoted as ${s_1},{s_2}, \cdots ,{s_n}$. The corresponding set of discipline embeddings is represented as ${\mathbb{S}_t} = \{ {\bm{s}_1},{\bm{s}_2}, \cdots ,{\bm{s}_n}\} $. The overall IC is calculated using the following formula:

\begin{equation}
    IC = \frac{{|{\mathbb{S}_t}| - |{\mathbb{S}_{\min }}|}}{{|{\mathbb{S}_{\max }}| - |{\mathbb{S}_{\min }}|}} + \frac{{\sum\limits_{i = 1}^{n - 1} {\sum\limits_{j = i + 1}^n {Dist({\bm{s}_i},{\bm{s}_j})} } }}{{C_{|{\mathbb{S}_t}|}^2}}
\end{equation}

where $\mathbb{S}_{\min }$ and $\mathbb{S}_{\max }$ represent the sets with the minimum and maximum number of disciplines in the dataset, respectively. ${C_{|{\mathbb{S}_t}|}^2}$ denotes the number of combinations of two elements selected from $|{\mathbb{S}_t}|$. ${Dist({\bm{s}_i},{\bm{s}_j})}$ represents the distance between discipline descriptions $s_i$ and $s_j$, defined by:

\begin{equation}
    Dist({s_i},{s_j}) = 1 - \frac{{{s_i} \cdot {s_j}}}{{\left\| {{s_i}} \right\|\left\| {{s_j}} \right\|}}
\end{equation}

\subsection{Extraneous Hardness Score}
We define the Extraneous Hardness Score as the mean of the Instruction-Response Expansion Index and the Silhouette Coefficient, characterizing the extrinsic difficulty of the data.
\subsubsection{Instruction-Response Expansion Index}

Instruction-Response Expansion Index (IREI) reflects the difficulty of data based on the lengths and ratio of the instruction and response. A longer combined length of the instruction and response indicates greater learning difficulty for the model. A higher ratio of response length to instruction length suggests that the instruction provides limited context, requiring the model to rely more on its internalized knowledge to generate an appropriate response. The specific calculation is defined as follows:
\begin{equation}
    IREI = \frac{{{L_{{\text{inst}}}} + {L_{{\text{resp}}}} - {L_{\min }}}}{{{L_{\max }} - {L_{\min }}}} + \frac{{{L_{{\text{resp}}}}}}{{{L_{{\text{inst}}}}}}
\end{equation}

where $L_\text{inst}$ and $L_\text{resp}$ denote the lengths of the instruction and the response, respectively, and ${{L_{\min }}}$ and ${{L_{\max }}}$ represent the minimum and maximum values of the combined instruction and response lengths across the dataset.
\subsubsection{Silhouette
Coefficient}
Another objective of the extraneous hardness score is to identify samples within the dataset that are both isolated and representative, as these are more likely to be unfamiliar to the model and therefore pose greater learning challenges. To this end, we conducted K-Means clustering on the dataset based on its TF-IDF vector representations \cite{bafna2016document}, and evaluated each sample using the silhouette coefficient:
\begin{equation}
    SC = \frac{{\alpha  - \beta }}{{\max \{ \alpha ,\beta \} }}
\end{equation}
where $\alpha$ denotes the average distance from the current sample to all samples in the nearest neighboring cluster, and $\beta$ represents the average distance from the current sample to all other samples within the same cluster.

\subsection{Overall Pipeline}

Finally, we construct a three-stage pipeline based on the previously introduced metrics. In the first stage, 80\% of the dataset is filtered out using the Reward Score. In the second stage, 50\% of the remaining samples are removed according to the Intrinsic Hardness Score. In the final stage, another 50\% of the data is filtered using the Extraneous Hardness Score, yielding the final training set.

\begin{table*}[htbp]  
\centering
\setlength{\arrayrulewidth}{0.3mm} 
\setlength{\tabcolsep}{6pt} 
\renewcommand{\arraystretch}{1.3} 
\resizebox{1\textwidth}{!}{ 
\begin{tabular}{c|cccccccccccccc|c}
\hline
\multirow{2}{*}{Method}& \multicolumn{15}{c}{\textbf{MMLU-Pro}} \\
\cline{2-16}
 &\textbf{Business}& \textbf{Law} & \textbf{Psychology} & \textbf{Biology} & \textbf{Chemistry} & \textbf{History} &\textbf{Health} &\textbf{Economics} &\textbf{Math} &\textbf{Physics}  &\textbf{Philosophy} &\textbf{Engineering} &\textbf{Computer Science}&\textbf{Other} & \textbf{Overall}  \\

\hline

Full-1B & 14.83 & 14.44 & 33.08 & 39.61 & 12.81 & 15.22 & 18.34 & 30.33 & 11.18 & 14.01 & 17.43 & 14.14 & 20.73 & 15.80 & 18.46  \\

Random Sampling-1B & 14.20 & 15.44 & 33.58 & 41.42 & \textbf{15.19} & \textbf{18.90} & 21.03 & 29.38 & 11.99 & 14.63 & 19.44 & 15.69 & 20.49 & 17.32 & 19.58  \\

AlpaGasus-1B & 14.20 & \textbf{17.17} & \textbf{35.34} & \textbf{46.58} & 15.02 & 17.59 & \textbf{22.13} & \textbf{31.75} & 13.47 & 14.93 & 18.24 & \textbf{17.23} & \textbf{22.20} & 17.21 & 20.67  \\

\rowcolor{mygray} Ours-1B & \textbf{21.17} & 15.44 & 30.58 & 41.42 & 14.13 & 17.85 & 21.39 & 29.38 & \textbf{22.35} & \textbf{15.55} & \textbf{20.04} & 14.34 & 20.24 & \textbf{19.05} & \textbf{21.04}  \\
\hline

Full-3B & 25.98 & 20.80 & 43.61 & 55.23 & 24.03 & 26.77 & 35.45 & 39.34 & 20.21 & 23.02 & 25.45 & 29.00 & 29.27 & 28.25 & 29.38  \\

Random Sampling-3B & 26.74 & \textbf{23.43} & 45.74 & 58.44 & 24.91 & \textbf{30.97} & 38.14 & \textbf{42.54} & 19.84 & 23.79 & \textbf{29.26} & \textbf{32.09} & 31.71 & 29.33 & 31.24  \\

AlpaGasus-3B & 29.78 & 20.98 & 42.73 & 55.23 & 26.50 & 27.56 & 35.09 & 39.70 & 23.61 & 24.17 & 29.06 & 30.55 & 28.29 & 29.98 & 30.73  \\

\rowcolor{mygray}Ours-3B& \textbf{33.59} & 20.98 & \textbf{48.37} & \textbf{61.51} & \textbf{28.36} & 29.66 & \textbf{39.36} & 42.42 & \textbf{25.54} & \textbf{26.02} & 26.65 & 30.75 & \textbf{32.20} & \textbf{31.49} & \textbf{33.03} \\ 
\hline

Full-8B & 27.25 & 26.34 & 48.99 & 60.53 & 22.70 & 35.17 & 44.50 & 42.77 & 20.13 & 23.09 & 33.27 & 32.30 & 34.39 & 32.90 & 32.76  \\

Random Sampling-8B & 30.80 & 28.97 & \textbf{54.26} & \textbf{64.30} & 27.56 & \textbf{38.58} & 50.49 & 48.22 & 23.98 & 28.64 & 36.67 & \textbf{36.02} & 36.34 & 36.90 & 37.01  \\

AlpaGasus-8B & 33.71 & \textbf{30.15} & 54.14 & 64.16 & 30.21 & 36.75 & 50.61 & \textbf{49.53} & 26.79 & 30.10 & 35.87 & 35.19 & 38.78 & \textbf{38.96} & 38.20  \\

\rowcolor{mygray}Ours-8B & \textbf{37.14} & 28.34 & 52.88 & 63.74 & \textbf{33.30} & 36.48 & \textbf{51.71} & 48.70 & \textbf{30.72} & \textbf{33.10} & \textbf{37.88} & 34.16 & \textbf{38.78} & 37.45 & \textbf{39.10} \\
\hline
\end{tabular}
}

\caption{Experiment Results on MMLU-Pro. The metrics in the table denote the model’s accuracy for each subject.}

\label{tab:mmlu-pro}
\end{table*}

\section{Experiment}
\subsection{Baselines and Experiment Setup}
\subsubsection{Baselines}
We compare our method against the following baselines:

\textbf{Full}: Refers to the Alpaca dataset \cite{taori2023alpaca} containing 52k instruction-following samples.

\textbf{Random Sampling}: Randomly select 2.6k samples from the Alpaca dataset.

\textbf{AlpaGasus} \cite{chenalpagasus}: A high-quality subset of Alpaca containing about 9.2k examples, from which we selected 2.6k samples for SFT.

\begin{table}[htbp]
\centering
\setlength{\arrayrulewidth}{0.3mm} 
\setlength{\tabcolsep}{6pt} 
\renewcommand{\arraystretch}{1.3} 
\resizebox{0.48\textwidth}{!}{ 
\begin{tabular}{c|ccccc|c}
\hline
\multirow{2}{*}{Method}& \multicolumn{6}{c}{\textbf{Alpaca-Eval}} \\
\cline{2-7}
 &\textbf{Helpful Base}& \textbf{Koala} & \textbf{Self-instruct} & \textbf{Oasst} & \textbf{Vicuna} & \textbf{Overall}  \\

\hline
Random Sampling-1B & 67.44 & 63.46 & 60.32 & 63.83 & 71.25 & 63.98  \\

AlpaGasus-1B & \textbf{85.27} & 72.44 & 71.03 & 76.60 & \textbf{78.75} & 75.65 \\

\rowcolor{mygray} Ours-1B & 79.84 & \textbf{76.92} & \textbf{73.02} & \textbf{79.26} & 75.00 & \textbf{76.52}  \\
\hline
Random Sampling-3B & 79.84 & 63.46 & 70.24 & \textbf{79.26} & 78.75 & 73.42 \\

AlpaGasus-3B & 77.52 & 59.62 & 69.05 &67.55  & 72.50 & 68.57   \\

 \rowcolor{mygray}Ours-3B& \textbf{87.60} & \textbf{77.56} & \textbf{75.79} & 78.72 & \textbf{85.00} & \textbf{79.63}  \\ 
\hline
Random Sampling-8B & 78.29 & 76.92 & 70.24 & 71.81 & 70.00 & 73.17 \\

AlpaGasus-8B & \textbf{92.25} & \textbf{83.33} & 74.21 & 83.51 & \textbf{87.50} & 82.36  \\

\rowcolor{mygray}Ours-8B & 87.60 & 82.05 & \textbf{80.56} & \textbf{85.11} & 81.25 & \textbf{83.11}  \\
\hline
\end{tabular}
}
\caption{On Alpaca-Eval, we report the win rate of each model over the full dataset baseline, judged by Qwen3-32B. To mitigate positional bias, response order is randomized during assessment.}

\label{tab:alpaca-eval}
\end{table}

\subsubsection{Experiment Setup}

We use Llama-3.2-1B/3B and Llama-3.1-8B-Instruct \cite{dubey2024llama} as base models, training all baselines with LoRA \cite{hu2022lora}, with consistent hyperparameter settings. Our method, consistent with the baselines, also uses 2.6k Alpaca samples \cite{taori2023alpaca} for training. Evaluation is conducted on AlpacaEval \cite{alpaca_eval} and MMLU-Pro \cite{wang2024mmlu}, with decoding temperature set to 0.

\subsection{Main Results}
We conducted a systematic evaluation of various data selection methods on two benchmarks—MMLU-Pro and Alpaca-Eval—focusing on their generalization performance after supervised fine-tuning (SFT). The experimental results are shown in Table \ref{tab:mmlu-pro} and Table \ref{tab:alpaca-eval}. Based on these results, we derive the following key observations:

First, on MMLU-Pro, models trained on the full dataset exhibited the lowest average performance across all model scales. On Alpaca-Eval, for all model sizes, every alternative method achieved a win rate above 50\% compared to the models trained on the full dataset. This suggests that a large quantity of unfiltered training data not only fails to effectively improve model capability, but may in fact harm generalization performance.

Second, Random Sampling, AlpaGasus, and our method all used only about 5\% of the original dataset for training, yet consistently outperformed the full-data models across all model sizes. This supports the “less is more” perspective: compared to large-scale but noisy datasets, a small amount of high-quality data can better enhance generalization.

Third, comparing the overall performance of Random Sampling and AlpaGasus reveals that data selection methods with explicit filtering criteria hold a clear advantage over purely random strategies. In most cases, AlpaGasus outperformed Random Sampling, indicating that large language models possess some discriminative ability in data selection. However, its reliance on internal model knowledge and the limited interpretability of its scoring mechanism constrain its potential for further improvement.

Finally, our method achieved the best overall performance across all model sizes: on MMLU-Pro, it reached accuracies of 21.04 (1B), 33.03 (3B), and 39.10 (8B); on Alpaca-Eval, the win rates over the full-dataset models were 76.52\% (1B), 79.63\% (3B), and 83.11\% (8B), respectively. Moreover, our method attained the best results in most subjects of MMLU-Pro, with particularly strong performance in high-cognitive-demand domains such as mathematics and physics. These findings indicate that, given the same data volume, our method prioritizes high-difficulty training samples can significantly enhance a model’s generalization to diverse tasks. This further validates our core hypothesis: higher-level cognitive and more difficult data is more valuable for model training.

\subsection{Annotation Guidance for Vertical-Domain Data}

\begin{table}[htbp]
\centering
\setlength{\arrayrulewidth}{0.3mm} 
\setlength{\tabcolsep}{6pt} 
\renewcommand{\arraystretch}{1.3} 
\resizebox{0.48\textwidth}{!}{ 
\begin{tabular}{c|c|c}
\hline
\textbf{Method} & \textbf{TCM Exam Test} & \textbf{THTB Score}\\

\hline
Vanilla & 55.15 & - \\

Simple Instruction Dataset & 58.79 & 0.3544 \\

Hard Instruction Dataset & \textbf{61.82}  & \textbf{0.5518} \\
\hline

\end{tabular}
}
\caption{Vanilla denotes Qwen3-8B. TCM Exam Test refers to the 2025 Chinese National Postgraduate Entrance Examination for Clinical Medicine (Traditional Chinese Medicine Comprehensive), consisting of 105 single-choice and 60 multiple-choice. The evaluation metric is accuracy.}

\label{tab:tcm}
\end{table}
To assess the utility of THTB in guiding data annotation for vertical domains, we conducted an empirical evaluation in the field of Traditional Chinese Medicine (TCM).

First, based on multiple textbooks in TCM gynecology, we used Qwen3-32B to construct 10k instruction samples as the baseline dataset. Then, under the guidance of THTB, we constructed 200 instruction samples with higher-level cognitive and greater difficulty using the same model. Finally, we separately performed supervised fine-tuning (SFT) of Qwen3-8B on these datasets.

The results are shown in Table \ref{tab:tcm}. The experiments indicate that the 200 high-difficulty instructions constructed under THTB guidance, despite being only 1/50 the size of the baseline dataset, achieved superior performance compared with the baseline. This finding validates the effectiveness of THTB in guiding data annotation. Furthermore, the three-stage average normalized scores computed using THTB show that the THTB-guided dataset has a higher mean hardness, validating the consistency of THTB in guiding data selection and annotation.

\section{Conclusion}
In this paper, we propose THTB, a framework that constructs hardness scores guided by Bloom’s Taxonomy to prioritize higher-level cognitive instructions, providing interpretable criteria for data selection and effective annotation guidance in vertical domains. Experiments show that THTB enables models trained on minimal data to outperform those trained on much larger datasets, demonstrating its effectiveness in both instruction data selection and annotation guidance.

\clearpage

\bibliographystyle{IEEEbib}
\bibliography{refs}

\begin{thebibliography}{10}

\bibitem{annepaka2025large}
Yadagiri Annepaka and Partha Pakray,
\newblock ``Large language models: A survey of their development, capabilities, and applications,''
\newblock {\em Knowledge and Information Systems}, vol. 67, no. 3, pp. 2967--3022, 2025.

\bibitem{zhang2024survey}
Bolin Zhang, Jiahao Wang, Qianlong Du, Jiajun Zhang, Zhiying Tu, and Dianhui Chu,
\newblock ``A survey on data selection for {LLM} instruction tuning,''
\newblock {\em arXiv preprint arXiv:2402.05123}, 2024.

\bibitem{wang2023self}
Yizhong Wang, Yeganeh Kordi, Swaroop Mishra, Alisa Liu, Noah~A Smith, Daniel Khashabi, and Hannaneh Hajishirzi,
\newblock ``{Self-Instruct}: Aligning language models with self-generated instructions,''
\newblock in {\em Proceedings of the 61st Annual Meeting of the Association for Computational Linguistics (Volume 1: Long Papers)}, 2023, pp. 13484--13508.

\bibitem{ding2023enhancing}
Ning Ding, Yulin Chen, Bokai Xu, Yujia Qin, Shengding Hu, Zhiyuan Liu, Maosong Sun, and Bowen Zhou,
\newblock ``Enhancing chat language models by scaling high-quality instructional conversations,''
\newblock in {\em Proceedings of the 2023 Conference on Empirical Methods in Natural Language Processing}, 2023, pp. 3029--3051.

\bibitem{zhou2023lima}
Chunting Zhou, Pengfei Liu, Puxin Xu, Srinivasan Iyer, Jiao Sun, Yuning Mao, Xuezhe Ma, Avia Efrat, Ping Yu, Lili Yu, et~al.,
\newblock ``{LIMA}: Less is more for alignment,''
\newblock {\em Advances in Neural Information Processing Systems}, vol. 36, pp. 55006--55021, 2023.

\bibitem{chenalpagasus}
Lichang Chen, Shiyang Li, Jun Yan, Hai Wang, Kalpa Gunaratna, Vikas Yadav, Zheng Tang, Vijay Srinivasan, Tianyi Zhou, Heng Huang, et~al.,
\newblock ``{AlpaGasus}: Training a better alpaca with fewer data,''
\newblock in {\em The Twelfth International Conference on Learning Representations}.

\bibitem{ge2024clustering}
Yuan Ge, Yilun Liu, Chi Hu, Weibin Meng, Shimin Tao, Xiaofeng Zhao, Mahong Xia, Zhang Li, Boxing Chen, Hao Yang, et~al.,
\newblock ``Clustering and ranking: Diversity-preserved instruction selection through expert-aligned quality estimation,''
\newblock in {\em Proceedings of the 2024 Conference on Empirical Methods in Natural Language Processing}, 2024, pp. 464--478.

\bibitem{li2024superfiltering}
Ming Li, Yong Zhang, Shwai He, Zhitao Li, Hongyu Zhao, Jianzong Wang, Ning Cheng, and Tianyi Zhou,
\newblock ``Superfiltering: Weak-to-strong data filtering for fast instruction-tuning,''
\newblock in {\em Proceedings of the 62nd Annual Meeting of the Association for Computational Linguistics (Volume 1: Long Papers)}, 2024, pp. 14255--14273.

\bibitem{anderson2001taxonomy}
Lorin~W Anderson and David~R Krathwohl,
\newblock {\em A taxonomy for learning, teaching, and assessing: A revision of {Bloom's} taxonomy of educational objectives: complete edition},
\newblock Addison Wesley Longman, Inc., 2001.

\bibitem{cai2024internlm2}
Zheng Cai, Maosong Cao, Haojiong Chen, Kai Chen, Keyu Chen, Xin Chen, Xun Chen, Zehui Chen, Zhi Chen, Pei Chu, et~al.,
\newblock ``{InternLM2} technical report,''
\newblock {\em arXiv preprint arXiv:2403.17297}, 2024.

\bibitem{yang2025qwen3}
An~Yang, Anfeng Li, Baosong Yang, Beichen Zhang, Binyuan Hui, Bo~Zheng, Bowen Yu, Chang Gao, Chengen Huang, Chenxu Lv, et~al.,
\newblock ``Qwen3 technical report,''
\newblock {\em arXiv preprint arXiv:2505.09388}, 2025.

\bibitem{xiao2024c}
Shitao Xiao, Zheng Liu, Peitian Zhang, Niklas Muennighoff, Defu Lian, and Jian-Yun Nie,
\newblock ``{C-Pack}: Packed resources for general chinese embeddings,''
\newblock in {\em Proceedings of the 47th international ACM SIGIR conference on research and development in information retrieval}, 2024, pp. 641--649.

\bibitem{bafna2016document}
Prafulla Bafna, Dhanya Pramod, and Anagha Vaidya,
\newblock ``Document clustering: {TF-IDF} approach,''
\newblock in {\em 2016 International conference on electrical, electronics, and optimization techniques (ICEEOT)}. IEEE, 2016, pp. 61--66.

\bibitem{taori2023alpaca}
Rohan Taori, Ishaan Gulrajani, Tianyi Zhang, Yann Dubois, Xuechen Li, Carlos Guestrin, Percy Liang, and Tatsunori~B Hashimoto,
\newblock ``Alpaca: A strong, replicable instruction-following model,''
\newblock {\em Stanford Center for Research on Foundation Models. https://crfm. stanford. edu/2023/03/13/alpaca. html}, vol. 3, no. 6, pp. 7, 2023.

\bibitem{dubey2024llama}
Abhimanyu Dubey, Abhinav Jauhri, Abhinav Pandey, Abhishek Kadian, Ahmad Al-Dahle, Aiesha Letman, Akhil Mathur, Alan Schelten, Amy Yang, Angela Fan, et~al.,
\newblock ``The {LLaMA} 3 herd of models,''
\newblock {\em arXiv e-prints}, pp. arXiv--2407, 2024.

\bibitem{hu2022lora}
Edward~J Hu, Yelong Shen, Phillip Wallis, Zeyuan Allen-Zhu, Yuanzhi Li, Shean Wang, Lu~Wang, Weizhu Chen, et~al.,
\newblock ``{LoRA}: Low-rank adaptation of large language models.,''
\newblock {\em ICLR}, vol. 1, no. 2, pp. 3, 2022.

\bibitem{alpaca_eval}
Xuechen Li, Tianyi Zhang, Yann Dubois, Rohan Taori, Ishaan Gulrajani, Carlos Guestrin, Percy Liang, and Tatsunori~B. Hashimoto,
\newblock ``{AlpacaEval}: An automatic evaluator of instruction-following models,'' \url{https://github.com/tatsu-lab/alpaca_eval}, 5 2023.

\bibitem{wang2024mmlu}
Yubo Wang, Xueguang Ma, Ge~Zhang, Yuansheng Ni, Abhranil Chandra, Shiguang Guo, Weiming Ren, Aaran Arulraj, Xuan He, Ziyan Jiang, et~al.,
\newblock ``{MMLU-Pro}: A more robust and challenging multi-task language understanding benchmark,''
\newblock {\em Advances in Neural Information Processing Systems}, vol. 37, pp. 95266--95290, 2024.

\end{thebibliography}

\end{document}